\documentclass[runningheads]{llncs}

\usepackage{amsmath,amssymb} 
\usepackage{color,comment,graphicx}

\usepackage[accsupp]{axessibility}  

\usepackage{pifont}
\usepackage[breaklinks,colorlinks,pagebackref]{hyperref}

\begin{document}

\pagestyle{headings}
\mainmatter
\def\ECCVSubNumber{6540}  

\title{TIPS: Text-Induced Pose Synthesis}

\titlerunning{TIPS: Text-Induced Pose Synthesis}

\author{Prasun Roy \inst{1} \and
Subhankar Ghosh \inst{1} \and
Saumik Bhattacharya \inst{2} \and\\
Umapada Pal \inst{3} \and
Michael Blumenstein \inst{1}}

\authorrunning{P. Roy et al.}

\institute{University of Technology Sydney, Australia\\
\email{prasun.roy@student.uts.edu.au, subhankar.ghosh@student.uts.edu.au, michael.blumenstein@uts.edu.au} \and
Indian Institute of Technology Kharagpur, India\\
\email{saumik@ece.iitkgp.ac.in} \and
Indian Statistical Institute Kolkata, India\\
\email{umapada@isical.ac.in}\\
\url{https://prasunroy.github.io/tips}}

\maketitle

\begin{abstract}
In computer vision, human pose synthesis and transfer deal with probabilistic image generation of a person in a previously unseen pose from an already available observation of that person. Though researchers have recently proposed several methods to achieve this task, most of these techniques derive the target pose directly from the desired target image on a specific dataset, making the underlying process challenging to apply in real-world scenarios as the generation of the target image is the actual aim. In this paper, we first present the shortcomings of current pose transfer algorithms and then propose a novel text-based pose transfer technique to address those issues. We divide the problem into three independent stages: (a) text to pose representation, (b) pose refinement, and (c) pose rendering. To the best of our knowledge, this is one of the first attempts to develop a text-based pose transfer framework where we also introduce a new dataset DF-PASS, by adding descriptive pose annotations for the images of the DeepFashion dataset. The proposed method generates promising results with significant qualitative and quantitative scores in our experiments.
\end{abstract}

\section{Introduction}\label{sec:introduction}

\begin{figure}[t]
  \centering
  \includegraphics[width=\linewidth]{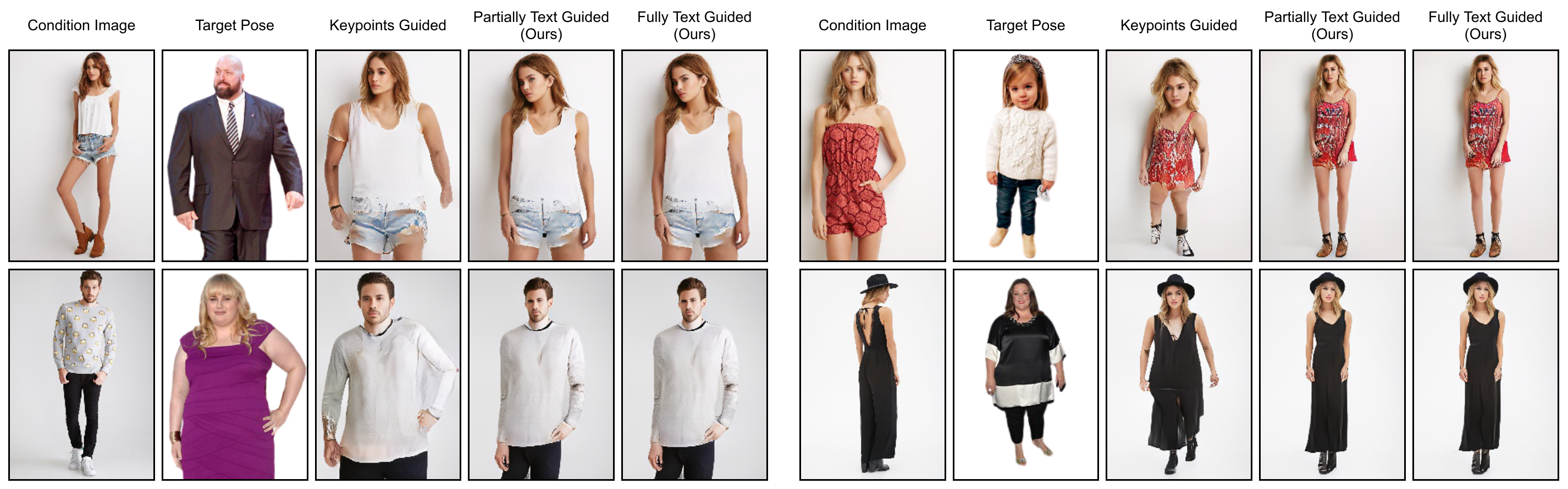}
  \caption{Overview of the proposed approach. Keypoint-guided methods tend to produce structurally inconsistent images when the physical appearance of the target pose reference significantly differs from the condition image. The proposed text-guided technique successfully addresses this issue while retaining the ability to generate visually decent results close to the keypoint-guided baseline.}
  \label{fig:introduction}
\end{figure}

Generating novel views of a given object is a challenging yet necessary task for many computer vision applications. Pose transfer is a subclass of the view synthesis problem where the goal is to estimate an unseen view (\textit{target} image) of a person with a particular pose from a given observation (\textit{source} image) of that person. As there can be significant differences between the source and target images, the pose transfer pipeline requires a very accurate generative algorithm to infer both the visible and occluded body parts in the target image. The method also needs to preserve the person's general appearance, including facial expression, skin color, attire, and background. In particular, the goal is to generate a target person image $I_B$ for a specific pose $P_B$ from an input source image $I_A$ of that person having an observed pose $P_A$. A human pose $P$ is usually expressed by a set of body-joint locations (\textit{keypoints}), denoted as $K$. As the location of the keypoints can vary significantly from person to person, two different sets of keypoints $K$ and $K'$ may represent the same pose $P$.

As initial solutions, researchers have introduced coarse to fine generation schemes \cite{ma2017pose,ma2018disentangled} by splitting the problem into separate sub-tasks for handling background, foreground, and pose separately. The architectural complexity of such an approach is later streamlined with a unified pipeline by utilizing deformable GANs \cite{siarohin2018deformable}, variational U-Net \cite{esser2018variational}, and progressive attention transfer \cite{zhu2019progressive}. Although the state-of-the-art (SOTA) algorithms have produced visually compelling results, a common yet noticeable flaw is present in these techniques. For training and evaluation of the models, SOTA algorithms extract keypoints $K_B$ directly from $I_B$ to represent $P_B$ and use it as one of the inputs. However, $I_B$ should not be ideally known to users, and such an over-simplified training process creates a dilemma. One way to circumvent the problem is training the model to adapt to a target pose $P_B$, represented by keypoints $K'_B$, which is extracted from the image $I'_B$ of some other person. However, as the models are trained using $K_B$ directly, they adapt poorly to any other set of keypoints $K'_B\neq K_B$ representing the same pose $P_B$. In Fig. \ref{fig:introduction}, we have shown the limitation of the existing keypoint-based models. We use the existing keypoint-guided pose transfer algorithm PATN \cite{zhu2019progressive} as a baseline in our experiments. The keypoint-based models try to follow the body structure of the target reference rather than the general pose. Thus, they fail occasionally in the absence of the target image $I_B$ to provide the keypoints. On the other hand, the proposed algorithm is not biased toward the target image as it exclusively works on the textual description of the target pose.

In this paper, we propose a novel pose-transfer pipeline guided by the textual description of the pose. Initially, we estimate the target keypoint set $K_B$ from the textual description $T_B$ of the target pose $P_B$. The estimated keypoint set $K_B$ is then used to generate the pose-transferred image $\tilde{I}_B$. As the estimation of $K_B$ is directly conditioned on $T_B$, we do not need the target image $I_B$ for estimating the target pose $P_B$, and the training is free from any bias. The main contributions of our work are as follows.
\begin{itemize}
    \item We propose a pose transfer pipeline that takes the source image and a textual description of the target pose to generate the target image. To the best of our knowledge, this is one of the first attempts to design a pose transfer algorithm based on textual descriptions of the pose.
    \item We introduce a new dataset DF-PASS derived from the DeepFashion dataset. The proposed dataset contains a human-annotated text description of the pose for 40488 images of the DeepFashion dataset.
    \item We extensively explore different perceptual metrics to analyze the performance of the proposed technique and introduce a new metric (GCR) for evaluating the gender consistency in the generated images.
    \item Most importantly, the algorithm is designed not to require the target image at the time of inference, making it more suitable for real-world applications than the existing pose transfer algorithms.
\end{itemize}

\section{Related Work}\label{sec:related_work}
Novel view synthesis is an intriguing problem in computer vision. Recently, Generative Adversarial Networks (GANs) \cite{goodfellow2014generative} have been explored extensively for perceptually realistic image generation \cite{goodfellow2014generative,johnson2016perceptual,lassner2017generative,ledig2017photo,mirza2014conditional,radford2016unsupervised}. Conditional generative models \cite{isola2017image,mirza2014conditional,sangkloy2017scribbler,zhu2017unpaired} have become popular in different fields of computer vision, such as inpainting \cite{yeh2017semantic}, super-resolution \cite{dong2015image,kim2016accurate} etc. Pose transfer can be viewed as a sub-category of the conditional generation task where a target image is generated from a source image by conditioning on the target pose. Thus, with the progress of conditional generative models, pose transfer algorithms have significantly enhanced performance in the last decade. Initial multi-stage approaches divide the complex task into relatively simpler sub-problems. In \cite{zhao2018multi}, Zhao \textit{et al.} adopt a coarse to fine approach to generate multi-view images of a person from a single observation. Ma \textit{et al.} \cite{ma2017pose,ma2018disentangled} introduce a multi-stage framework to generate the final pose-transferred image from a single source image. Balakrishnan \textit{et al.} \cite{balakrishnan2018synthesizing} propose a method of pose transfer by segmenting and generating the foreground and background individually. Wang \textit{et al.} \cite{wang2018toward} introduce a characteristic preserving generative network with a geometric matching module. The coarse to fine generation technique is further improved by incorporating the idea of disentanglement \cite{ma2018disentangled} where the generative model is designed as a multi-branch network to handle foreground, background, and pose separately. In \cite{pumarola2018unsupervised}, the authors propose a pose conditioned bidirectional generator in an unsupervised multi-level generation strategy. In \cite{zhu2019progressive}, the authors introduce a progressive attention transfer technique to transfer the pose gradually. Li \textit{et al.} \cite{li2020pona} propose a method to progressively select important regions of an image using pose-guided non-local attention with a long-range dependency. Researchers have also investigated 3D appearance flow \cite{li2019dense}, pose flow \cite{zheng2020pose}, and surface-based modeling \cite{guler2018densepose,neverova2018dense} for pose transfer. In \cite{zanfir2018human}, the authors first approximate a 3D mesh from a single image, and then the 3D mesh is used to transfer the pose. Siarohin \textit{et al.} \cite{siarohin2018deformable} propose a nearest neighbour loss for pose transfer using deformable GANs. In \cite{chen2019hierarchical,zhou2020makeittalk}, the authors generate talking-face dynamics from a single face image and a target audio signal.

Text-based image generations are also an intriguing topic in computer vision. In \cite{reed2016generative}, the authors propose a GAN-based architecture for synthesizing the images. Qiao \textit{et al.} \cite{qiao2019mirrorgan} have incorporated redescription of textual descriptions for image synthesis. Recently, text-based approaches are also explored for generating human pose \cite{briq2021towards} and appearance \cite{zhang2020adversarial}. In \cite{li2018video}, the authors use a Variational Autoencoder (VAE) to generate human actions from text descriptions. In \cite{briq2021towards}, the authors generate 3D human meshes from text using a recurrent GAN and SMPL \cite{loper2015smpl} model. In \cite{zhou2019text}, the authors propose a text-guided method for generating human images by selecting a pose from a set of eight basic poses, followed by controlling the appearance attributes of the selected basic pose. However, text-based visual generation techniques are limited in the literature, and text-guided pose transfer is not well-explored previously to the best of our knowledge.

\section{Methodology}\label{sec:methodology}

\begin{figure}[t]
  \centering
  \includegraphics[width=\linewidth]{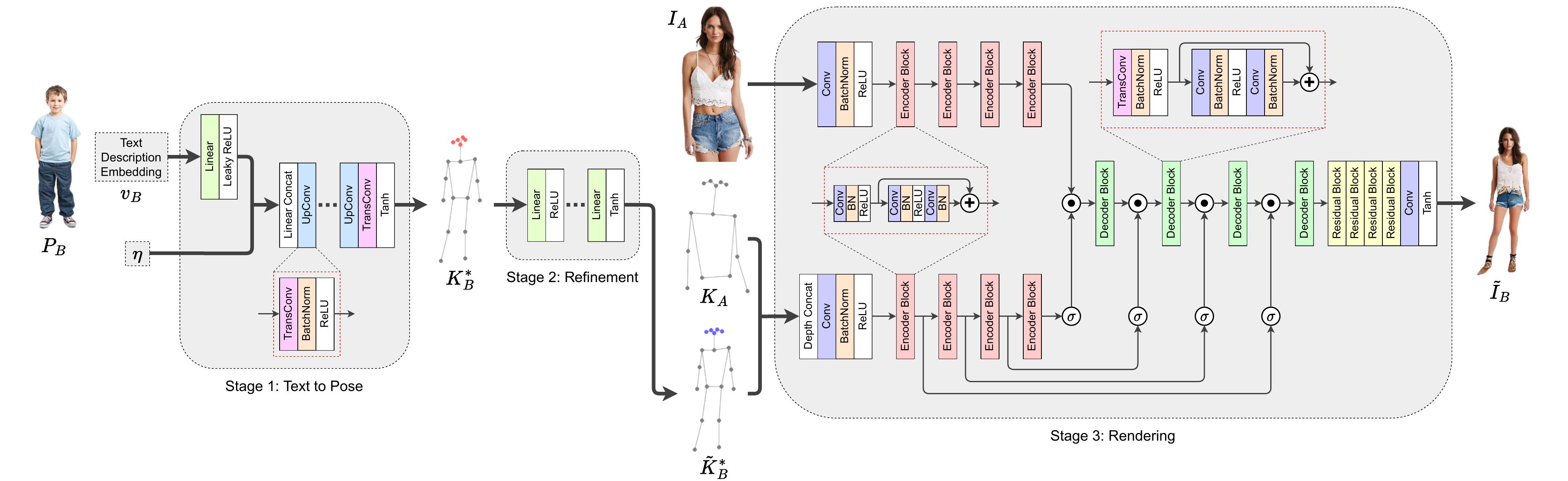}
  \caption{Architecture of the proposed pipeline. The workflow is divided into three stages. In stage 1, we estimate a spatial representation $K^*_B$ for the target pose $P_B$ from the corresponding text description embedding $v_B$. In stage 2, we regressively refine the initial estimation of the facial keypoints to obtain the refined target keypoints $\tilde{K}^*_B$. Finally, in stage 3, we render the target image $\tilde{I}_B$ by conditioning the pose transfer on the source image $I_A$ having the keypoints $K_A$ corresponding to the source pose $P_A$.}
  \label{fig:architecture}
\end{figure}

The proposed technique is divided into three independent sequential stages, each specific to a particular task. In the first stage, we derive an initial estimation of the target pose from the corresponding text description embedding. This coarse pose is then refined through regression at the next step. Finally, pose transfer is performed by conditioning the transformation on the appearance of the source image. We show our integrated generation pipeline in Fig. \ref{fig:architecture}.

\subsection{Text to Keypoints Generation}
For a given source image $I_A$, our algorithm aims to generate the pose-transferred image $I_B$ where the target pose $P_B$ is described by textual description $T_B$. At first, we encode $T_B$ into an embedded vector $v_B$ either by many-hot encoding or using a pre-trained NLP model such as BERT \cite{devlin2018bert}, FastText \cite{athiwaratkun2018probabilistic}, or Word2Vec \cite{mikolov2013efficient}. We first aim to estimate the keypoint set $K_B$ from $v_B$ using a generative model to guide the pose transfer process in a later stage. To train such a generative model, we represent the keypoints $k_j \in \mathbb{R}^{m \times n}$ where $k_j \in K; \forall j$ and the domain of both $I_A$ and $I_B$ is $\mathbb{R}^{m \times n}$. As a slight spatial variation of $k_j$ does not change the pose $P_B$, it is better to represent it with a Gaussian distribution $\mathcal{N}(k_j, \sigma_j)$; $\forall j$ for mitigating the high sparsity in the data. Although for different $k_j$, the invariance of the pose is valid for different amounts of spatial perturbations, we can assume $\sigma_j = \sigma$, a constant, $\forall j$, if $\sigma_j$ is small. Such representation of keypoints is often referred to as heatmaps.

Taking motivation from \cite{zhang2020adversarial}, we design a generative adversarial network to estimate the target keypoint set $K_B$ from the text embedding $v_B$. In our generator $G_T$, we first project $v_B$ into a 128-dimensional latent space $\phi_B$ using a linear layer with leaky ReLU activation. To allow some structural variations in the generated poses, we sample a 128-dimensional noise vector $\eta \sim \textit{N}(\mathbf{0}, \textit{I})$, where \textit{I} is a $128 \times 128$ identity matrix. Both $\phi_B$ and $\eta$ are linearly concatenated and passed through 4 up-convolution blocks. At each block, we perform a transposed convolution followed by batch normalization \cite{ioffe2015batch} and ReLU activation \cite{nair2010rectified}. The four transposed convolutions use 256, 128, 64, and 32 filters, respectively. We produce the final output from $G_T$ by passing the output of the last up-convolution block through another transposed convolution layer with 18 filters and $tanh$ activation. The final generator output $G_T(v_B, \eta)$ has a spatial dimension of $64 \times 64 \times 18$, where each channel represents one of the 18 keypoints $k_j$, $j \in \{1, 2, \hdots, 18\}$. In our discriminator (\textit{critic}) $D_T$, we first perform 4 successive convolutions, each followed by leaky ReLU activation, on the 18-channel heatmap. The four convolutions use 32, 64, 128, and 256 filters, respectively. The output of the last convolution layer is concatenated with 16 copies of $\phi_B$ arranged in a $4 \times 4$ tile. The concatenated feature map is then passed through a point convolution layer with 256 filters and leaky ReLU activation. We estimate the final scalar output from $D_T$ by passing the feature map through another convolution layer with a single filter. We mathematically define the objective function for $D_T$ as
\begin{equation}
    L_D = -\mathbb{E}_{(x, v_B) \sim p_t, \eta \sim p_\eta} [D_T(x, v_B) - D_T(G_T(\eta, v_B), v_B)]
    \label{eq:L_D}
\end{equation}
where $(x, v_B) \sim p_t$ is the heatmap and text embedding pair sampled from the training set, $\eta \sim p_\eta$ is the noise vector sampled from a Gaussian distribution, and $G_T(\eta, v_B)$ is the generated heatmap for the given text embedding $v_B$.
Researchers \cite{gulrajani2017improved} have shown that the WGAN training is more stable if $D_T$ is Lipschitz continuous, which mitigates the undesired behavior due to gradient clipping. To enforce the Lipschitz constraint, we compute gradient penalty as
\begin{equation}
    \mathcal{C}_T = \mathbb{E}_{(\tilde{x}, v_B) \sim p_{\tilde{x}, v_B}} [(\|\nabla_{\tilde{x}, v_B} D_T(\tilde{x}, v_B)\|_2 - 1) ^ 2]
    \label{eq:GP}
\end{equation}
where $\|.\|_2$ indicates the $l_2$ norm and $\tilde{x}$ is an interpolated sample between a real sample $x$ and a generated sample $G_T(\eta, v_B)$, i.e., $\tilde{x} = \alpha x + (1 - \alpha) G_T(\eta, v_B)$, where $\alpha$ is a random number, selected from a uniform distribution between 0 and 1. Eq. \ref{eq:GP} enforces the Lipschitz constraint by restricting the gradient magnitude to 1.
We define the overall objective of $D_T$ by combining equations \ref{eq:L_D} and \ref{eq:GP} as
\begin{equation}
    L_{D_T} = L_D + \lambda \mathcal{C}_T
    \label{eq:L_D_T}
\end{equation}
where $\lambda$ is a regularization constant. We keep $\lambda = 10$ in all of our experiments. We mathematically define the objective function for $G_T$ as
\begin{equation}
    \begin{split}
        L_{G_T} =& -\mathbb{E}_{\eta \sim p_\eta, v_B \sim p_{v_B}} [D_T(G_T(\eta, v_B), v_B)]\\
                 & -\mathbb{E}_{\eta \sim p_\eta, v^1_B, v^2_B \sim p_{v_B}} \left[D_T(G_T(\eta, \frac{v^1_B + v^2_B}{2}), \frac{v^1_B + v^2_B}{2})\right]
    \end{split}
    \label{eq:L_G_T}
\end{equation}
where $v^1_B, v^2_B \sim p_{v_B}$ are text encodings sampled from the training set. The second term in Eq. \ref{eq:L_G_T} helps the generator learn from the interpolated text encodings, which are not originally present in the training set.

We estimate the target keypoint set $K^*_B$ from the 18-channel heatmap generated from $G_T$ by computing the maximum activation $\psi^{max}_j$, $j \in \{1, 2, \hdots, 18\}$ for every channel. The spatial location of the maximum activation for the $j$-th channel determines the coordinates of the $j$-th keypoint if $\psi^{max}_j \geq 0.2$. Otherwise, the $j$-th keypoint is considered occluded if $\psi^{max}_j < 0.2$.

\subsection{Facial Keypoints Refinement}
While $G_T$ produces a reasonable estimate of the target keypoints from the corresponding textual description, the estimation $K^*_B$ is often noisy. The spatial perturbation is most prominent for the facial keypoints (nose, two eyes, and two ears) due to their proximity. Slight positional variations for other keypoints generally do not drastically affect the pose representation. Therefore, we refine the initial estimate of the facial keypoints by regression using a linear fully-connected network $N_R$ (RefineNet). At first, the five facial keypoints $k^f_i$, $i \in \{1, 2, \hdots, 5\}$ are translated by $(k^f_i - k_n)$ where $k_n$ is the spatial location of the nose. In this way, we align the nose with the origin of the coordinate system. Then, we normalize the translated facial keypoints such that the scaled keypoints $k^s_i$ are within a square of span $\pm 1$ and the scaled nose is at the origin (0, 0). Next, we flatten the coordinates of the five normalized keypoints to a 10-dimensional vector $v_f$ and pass it through three linear fully-connected layers, where each layer has 128 nodes and ReLU activation. The final output layer of the network consists of 10 nodes and $tanh$ activation. While training, we augment $k^s_i$ with small amounts of random 2D spatial perturbations and try to predict the original values of $k^s_i$. We optimize the parameters of $N_R$ by minimizing the mean squared error (MSE) between the actual and the predicted coordinates. Finally, we denormalize and retranslate the predicted facial keypoints. The refined set of keypoints $\tilde{K}^*_B$ is obtained by updating the coordinates of the facial keypoints of $K^*_B$ with the predictions from RefineNet.

\subsection{Pose Rendering} \label{sec:pose_transfer}
To render the final pose-transferred image $\tilde{I}_B$, we first extract the keypoints $K_A$ from the source image $I_A$ using a pre-trained Human Pose Estimator (HPE) \cite{cao2017realtime}. However, we may also estimate the keypoints $K^*_A$ from the embedding vector $v_A$ for the text description $T_A$ of the source pose $P_A$. If we compute the keypoints $K^*_A$ from $T_A$, then the refinement is also applied on $K^*_A$ to obtain the refined source keypoints $\tilde{K}^*_A$. Thus, depending on the source keypoints selection, we propose two slightly different variants of the method -- (a) partially text-guided, where we use HPE to extract $K_A$ and (b) fully text-guided, where we estimate $\tilde{K}^*_A$ using $G_T$ followed by $N_R$. However, in both cases, the pose rendering step works similarly. For simplicity, we discuss the rendering network using $K_A$ as the notation for the source keypoints. We represent the keypoints $K_A$ and $\tilde{K}^*_B$ as multi-channel heatmaps $H_A$ and $\tilde{H}_B$, respectively, where each channel of a heatmap corresponds to one particular keypoint.

We adopt an attention-guided conditional GAN architecture \cite{roy2022multi,roy2022scene} for the target pose rendering. We take $I_A$, $H_A$, and $\tilde{H}_B$ as inputs for our generator network $G_S$, which produces the final rendered image output $\tilde{I}_B$ as an estimate for the target image $I_B$. The discriminator network $D_S$ utilizes a PatchGAN \cite{isola2017image} to evaluate the quality of the generated image by taking a channel-wise concatenation between $I_A$ and either $I_B$ or $\tilde{I}_B$. In $G_S$, we have two downstream branches for separately encoding the condition image $I_A$ and the channel-wise concatenated heatmaps $(H_A, \tilde{H}_B)$. After mapping both inputs to a $256 \times 256$ feature space by convolution (kernel size = $3 \times 3$, stride = 1, padding = 1, bias = 0), batch normalization, and ReLU activation, we pass the feature maps through four consecutive encoder blocks. Each block encodes the input feature space by reducing the dimension to half but doubling the number of filters. Each encoder block features a sequence of convolution (kernel size = $4 \times 4$, stride = 2, padding = 1, bias = 0), batch normalization, ReLU activation, and a basic residual block \cite{he2016deep}. We combine the encoded feature maps and pass the merged feature space through an upstream branch with four consecutive decoder blocks. Each block decodes the feature space by doubling the dimension but reducing the number of filters by half. Each decoder block features a sequence of transposed convolution (kernel size = $4 \times 4$, stride = 2, padding = 1, bias = 0), batch normalization, ReLU activation, and a basic residual block. We use attention links between encoding and decoding paths at every resolution level to retain coarse and fine attributes in the generated image. Mathematically, for the lowest resolution level, $L = 4$,
\begin{equation*}
  I^\delta_3 = \delta_4 (I^{\pi^i}_4 \;\odot\; \sigma(H^{\pi^h}_4))
\end{equation*}
and for the higher resolution levels, $L = \{1, 2, 3\}$,
\begin{equation*}
  I^\delta_{L-1} = \delta_L (I^{\pi^i}_L \;\odot\; \sigma(H^{\pi^h}_L))
\end{equation*}
where, at the resolution level L, $I^\delta_L$ denotes the output of the decoding block $\delta_L$, $I^{\pi^i}_L$ denotes the output of the image encoding block $\pi^i_L$, $H^{\pi^h}_L$ denotes the output of the pose encoding block $\pi^h_L$, $\sigma$ is an element-wise \emph{sigmoid} activation function, and $\odot$ is an element-wise product. Finally, we pass the resulting feature maps through four consecutive basic residual blocks followed by a point-wise convolution (kernel size = $1 \times 1$, stride = 1, padding = 0, bias = 0) with \emph{tanh} activation to map the feature space into a $256 \times 256 \times 3$ normalized image $\tilde{I}_B$.

The optimization objective of $G_S$ consists of three loss components -- a pixel-wise $l_1$ loss $\mathcal{L}^{G_S}_{l_1}$, a discrimination loss $\mathcal{L}^{G_S}_{GAN}$ by $D_S$, and a perceptual loss $\mathcal{L}^{G_S}_{P_\rho}$ computed using a pre-trained VGG-19 network \cite{simonyan2015very}. We measure the pixel-wise $l_1$ loss as $\mathcal{L}^{G_S}_{l_1} = \|\tilde{I}_B - I_B\|_1$, where $\|.\|_1$ denotes the $l_1$ norm or the mean absolute error. We compute the discrimination loss as
\begin{equation}
  \mathcal{L}^{G_S}_{GAN} = \mathcal{L}_{BCE}(D_S(I_A, \tilde{I}_B), 1)
\end{equation}
where $\mathcal{L}_{BCE}$ denotes the binary cross-entropy loss. Finally, we estimate the perceptual loss as
\begin{equation}
  \mathcal{L}^{G_S}_{P_\rho} = \frac{1}{h_\rho w_\rho c_\rho} \sum_{x=1}^{h_\rho} \sum_{y=1}^{w_\rho} \sum_{z=1}^{c_\rho} \|q_\rho(\tilde{I}_B) - q_\rho(I_B)\|_1
\end{equation}
where $q_\rho$ is the output of dimension $(h_\rho \times w_\rho \times c_\rho)$ from the $\rho$-th layer of a pre-trained VGG-19 network. We add two perceptual loss terms for $\rho = 4$ and $\rho = 9$ to the objective function. So, in our method, the overall optimization objective for $G_S$ is given by
\begin{equation}
  \mathcal{L}_{G_S} = \lambda_1 \mathcal{L}^{G_S}_{l_1} + \lambda_2 \mathcal{L}^{G_S}_{GAN} + \lambda_3 (\mathcal{L}^{G_S}_{P_4} + \mathcal{L}^{G_S}_{P_9})
\end{equation}
where $\lambda_1$, $\lambda_2$, and $\lambda_3$ denote the weighing parameters for respective loss terms. We keep $\lambda_1 = 5$, $\lambda_2 = 1$, and $\lambda_3 = 5$ in our experiments. Lastly, we define the optimization objective for $D_S$ as
\begin{equation}
  \mathcal{L}_{D_S} = \frac{1}{2} \left[\mathcal{L}_{BCE}(D_S(I_A, I_B), 1) + \mathcal{L}_{BCE}(D_S(I_A,\tilde{I}_B), 0)\right]
\end{equation}

\section{Dataset and Training}\label{sec:data_training}
As this is one of the earliest attempts to perform a text-guided pose transfer, we introduce a new dataset called \emph{DeepFashion Pose Annotations and Semantics} (DF-PASS) to compensate for the lack of similar public datasets. DF-PASS contains a human-annotated textual description of the pose for 40488 images of the DeepFashion dataset \cite{liu2016deepfashion}. Each text annotation contains (1) the person's gender (e.g. `man', `woman' etc.); (2) visibility flags of the body keypoints (e.g. `his left eye is visible', `her right ear is occluded' etc.); (3) head and face orientations (e.g. `her head is facing partially left', `he is keeping his face straight' etc.); (4) body orientation (e.g. `facing towards front', `facing towards right' etc.); (5) hand and wrist positioning (e.g. `his right hand is folded', `she is keeping her left wrist near left hip' etc.); (6) leg positioning (e.g. `both of his legs are straight', `her right leg is folded' etc.). We recruit five in-house annotators to acquire the text descriptions, which two independent verifiers have validated. Each annotator describes a pose during data acquisition by selecting options from a set of possible attribute states. In this way, we have collected many-hot embedding vectors alongside the text descriptions. We use 37344 samples for training and 3144 samples for testing out of 40488 annotated samples following the same data split provided by \cite{zhu2019progressive}.

In stage 1, the text to pose conversion network uses the stochastic Adam optimizer \cite{kingma2015adam} to train both $G_T$ and $D_T$. We keep learning rate $\eta_1 = 1e^{-4}$, $\beta_1 = 0$, $\beta_2 = 0.9$, $\epsilon = 1e^{-8}$, and weight decay = 0 for the optimizer. While training, we update $G_T$ once after every five updates of $D_T$. In stage 2, we train the facial keypoints refinement network $N_R$ using stochastic gradient descent keeping learning rate $\eta_2 = 1e^{-2}$. In stage 3, the pose rendering network also uses the Adam optimizer to train both $G_S$ and $D_S$. In this case, we keep learning rate $\eta_3 = 1e^{-3}$, $\beta_1 = 0.5$, $\beta_2 = 0.999$, $\epsilon = 1e^{-8}$, and weight decay = 0. Before training, the parameters of $G_T$, $D_T$, $G_S$, and $D_S$ are initialized by sampling from a normal distribution of 0 mean and 0.02 standard deviation. The code is available at \url{https://github.com/prasunroy/tips}.

\section{Results}\label{sec:results}

\begin{figure}[t]
  \centering
  \includegraphics[width=\linewidth]{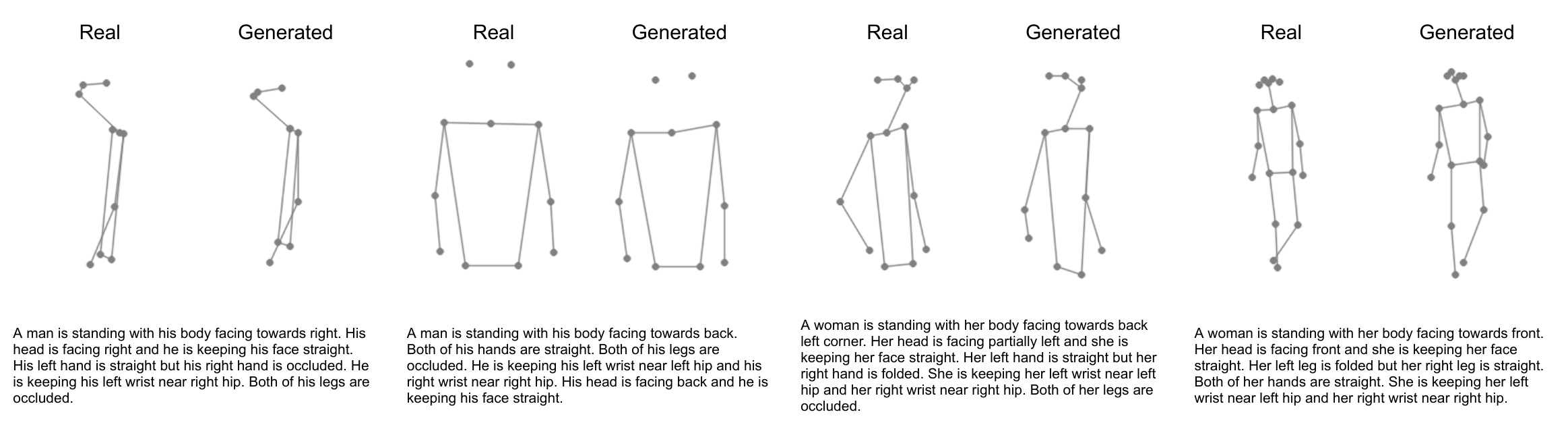}
  \caption{Qualitative results of text to pose generation using $G_T$.}
  \label{fig:result_stage1}
\end{figure}

\begin{figure}[t]
  \centering
  \includegraphics[width=0.8\linewidth]{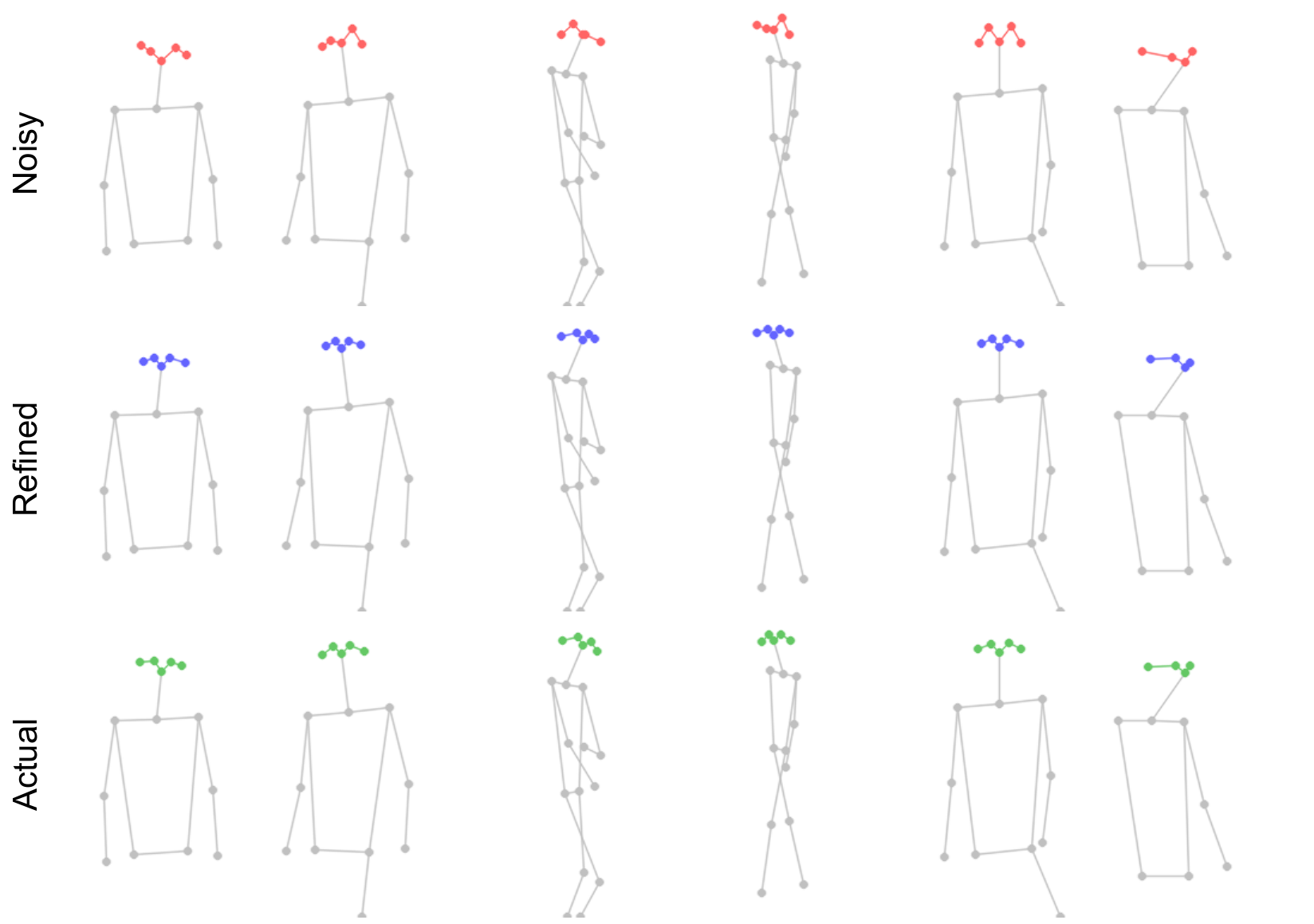}
  \caption{Qualitative results of regressive refinement using $N_R$.}
  \label{fig:result_stage2}
\end{figure}

\begin{figure}[t]
  \centering
  \includegraphics[width=\linewidth]{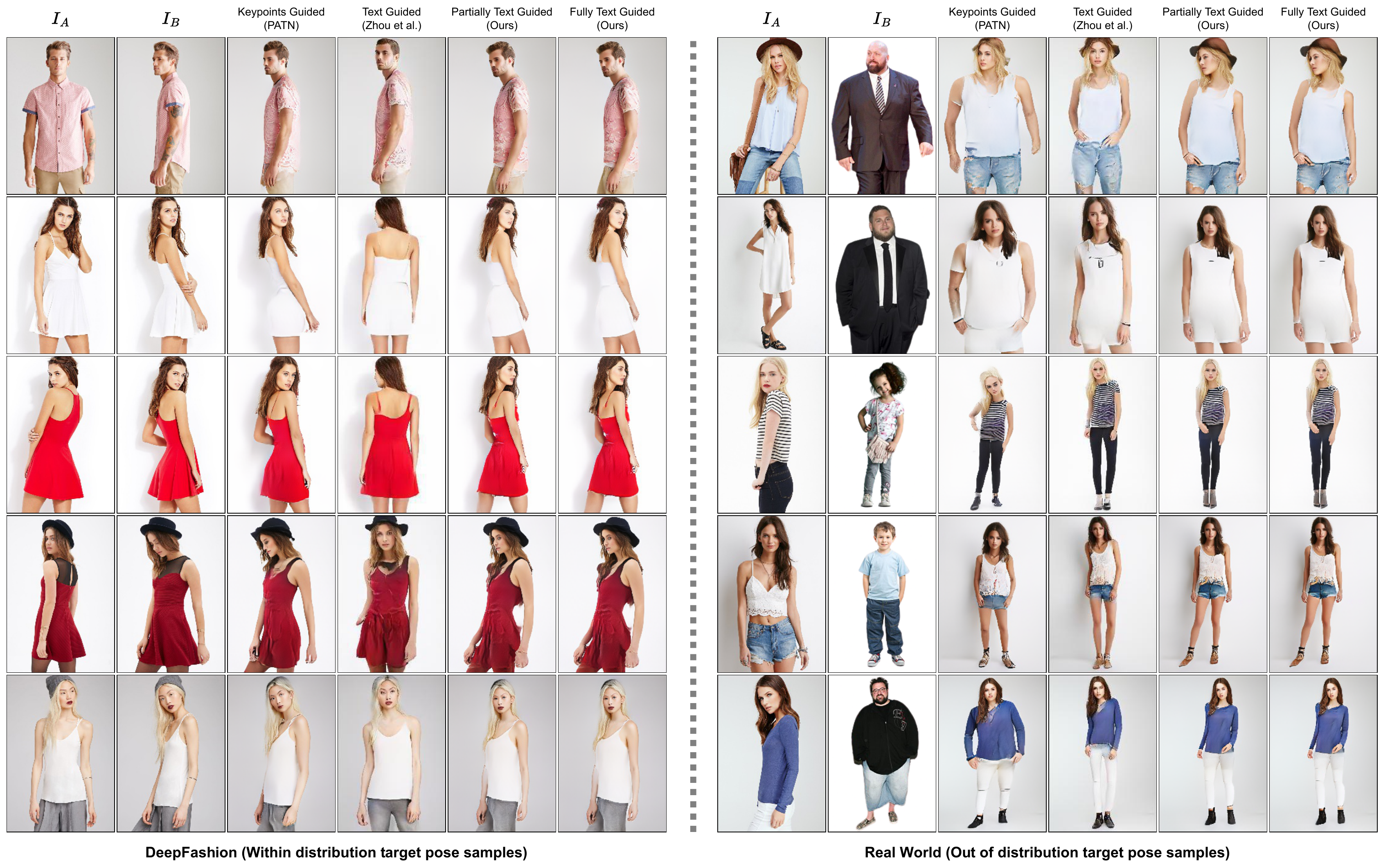}
  \caption{Qualitative results of different pose transfer algorithms.}
  \label{fig:result_stage3}
\end{figure}

In Fig. \ref{fig:result_stage1}, we demonstrate the output $K^*_B$ of the text to keypoints generator $G_T$. The textual descriptions used for estimating the respective $K^*_B$ are also shown in the figure. It can be observed that the estimated keypoints $K^*_B$ capture the pose $P_B$ and closely resembles $K_B$. However, a precise observation may reveal that the facial keypoints of $K^*_B$ significantly differ from $K_B$. In Fig. \ref{fig:result_stage2}, the advantage of regressive refinement of keypoints $K^*_B$ is shown. As depicted in the figure, the refinement network aims to rectify only the facial keypoints. In Fig. \ref{fig:result_stage3}, we demonstrate that when $K_B$ is selected from the DeepFashion dataset, the existing PATN algorithm performs satisfactorily. However, when $K_B$ comes from real-world samples (out of the DeepFashion dataset), PATN fails to generate the pose-transferred images while maintaining structural consistency. In the case of the proposed algorithm, as the pose description does not require structural information of the target image, the generated pose-transferred images are consistent with the respective source images. Our proposed algorithm performs well irrespective of the representation of the source pose, i.e., for the partially text-guided approach, where the source pose is represented using keypoints, and for the fully text-guided approach, where the source pose is described using text.

\subsection{Evaluation}
As the proposed method has three major steps -- text to keypoint generation, refinement of the generated keypoints, and generation of the pose-transferred image, it is important to analyze each step qualitatively and quantitatively.\\
\textbf{Metrics:} Quantifying the generated image quality is a challenging problem. However, researchers \cite{esser2018variational,ma2017pose,siarohin2018deformable,zhu2019progressive} have used a few well-known quantitative metrics to judge the quality of the synthesis. This includes a Structural Similarity Index (SSIM) \cite{wang2004image}, Inception Score (IS) \cite{salimans2016improved}, Detection Score (DS) \cite{liu2016ssd}, and PCKh \cite{andriluka20142d}. We also evaluate the Learned Perceptual Image Patch Similarity (LPIPS) \cite{zhang2018unreasonable} metric as a more modern replacement of the SSIM for perceptual image quality assessment. In our evaluation, we calculate LPIPS using both VGG19 \cite{simonyan2015very} and SqueezeNet \cite{iandola2016squeezenet} backbones. As we are dealing with human poses and evaluating the generation quality, we propose a novel metric, named Gender Consistency Rate (GCR), that evaluates whether the generated image $\tilde{I}_B$ can be identified to be of the same gender as the source image $I_A$ by a pre-trained classifier. GCR serves two purposes: first, it ensures that the gender-specific features are present in the generated image, and second, it ensures that the generated target image is consistent with the source image. To calculate GCR, we remove the last layer of the VGG19 network and add a single neuron with sigmoid activation to design a binary classifier and train it with the image samples from the DeepFashion dataset with label 0 for males and label 1 for females. The pre-trained network achieves a test accuracy of 0.995. We use this pre-trained model to compute the gender recognition rate for the generated images.\\
In Table \ref{tab:ablation_1}, we evaluate our proposed algorithm on the DeepFashion dataset. The keypoints-guided baseline \cite{zhu2019progressive} performs well for \emph{within distribution} target poses from DeepFashion. However, the proposed text-guided approach performs satisfactorily, as reflected in SSIM, IS, DS, and LPIPS scores. As PCKh uses keypoint coordinates, our method achieves a low PCKh score compared to the keypoint-based method, which uses precise keypoints for the target image generation. For evaluating \emph{out of distribution} target poses, we select 50 pairs of source and target images from DeepFashion; however, we estimate the target keypoints from real-world images (outside DeepFashion) having similar poses as the original target images. As shown in Table \ref{tab:result_realworld}, in such a case, the proposed technique achieves significantly higher SSIM and PCKh values, indicating much better structural generation for real-world pose references.\\
\textbf{User Study:} It is known that quantitative metrics do not always reflect the perceptual quality of images well \cite{siarohin2018deformable,zhu2019progressive} and a quantifiable metric for evaluating image quality is still an open problem in computer vision. Therefore, we also perform an opinion-based user assessment to judge the realness of the generated images. Following the similar protocol as \cite{ma2017pose,siarohin2018deformable,zhu2019progressive}, the observer needs to provide an instant decision whether an image is real or fake. We create a subset of 260 real and 260 generated images with 10 images of each type used as a practice set. During the test, 20 random images (10 real + 10 fake) are drawn from the remaining images and shown to the examiner. We compute the \textbf{R2G} (the fraction of real images identified as generated) and \textbf{G2R} (the fraction of generated images identified as real) scores from the user submissions. Our method achieves a mean G2R score of 0.6968 for submissions by 156 individual volunteers.

\begin{table}[t]
\centering
\caption{Performance of pose transfer algorithms on DeepFashion.}
\label{tab:ablation_1}
\resizebox{1.0\linewidth}{!}{%
\begin{tabular}{l|ccccccc}
\hline
\textbf{Pose Generation Algorithm} &
  \textbf{SSIM} &
  \textbf{IS} &
  \textbf{DS} &
  \textbf{PCKh} &
  \textbf{GCR} &
  \textbf{\begin{tabular}[c]{@{}c@{}}LPIPS\\ (VGG)\end{tabular}} &
  \textbf{\begin{tabular}[c]{@{}c@{}}LPIPS\\ (SqzNet)\end{tabular}} \\ \hline
Partially Text Guided (Ours)    & 0.549 & 3.269 & 0.950 & 0.53 & 0.963 & 0.402 & 0.290 \\
Fully Text Guided (Ours)        & 0.549 & 3.296 & 0.950 & 0.53 & 0.963 & 0.402 & 0.289 \\
Zhou et al. \cite{zhou2019text} & 0.373 & 2.320 & 0.864 & 0.62 & 0.979 & 0.310 & 0.215 \\
PATN \cite{zhu2019progressive}  & 0.773 & 3.209 & 0.976 & 0.96 & 0.983 & 0.299 & 0.170 \\ \hline
Real Data                       & 1.000 & 3.790 & 0.948 & 1.00 & 0.995 & 0.000 & 0.000 \\ \hline
\end{tabular}%
}
\end{table}

\begin{table}[t]
\centering
\caption{Performance of pose transfer algorithms for real-world targets.}
\label{tab:result_realworld}
\resizebox{1.0\linewidth}{!}{%
\begin{tabular}{l|ccccccc}
\hline
\textbf{Pose Generation Algorithm} &
  \textbf{SSIM} &
  \textbf{IS} &
  \textbf{DS} &
  \textbf{PCKh} &
  \textbf{GCR} &
  \textbf{\begin{tabular}[c]{@{}c@{}}LPIPS\\ (VGG)\end{tabular}} &
  \textbf{\begin{tabular}[c]{@{}c@{}}LPIPS\\ (SqzNet)\end{tabular}} \\ \hline
Partially Text Guided (Ours)    & 0.696 & 2.093 & 0.990 & 0.84 & 1.000 & 0.262 & 0.155 \\
Fully Text Guided (Ours)        & 0.695 & 2.171 & 0.991 & 0.85 & 1.000 & 0.263 & 0.157 \\
Zhou et al. \cite{zhou2019text} & 0.615 & 2.891 & 0.931 & 0.52 & 1.000 & 0.271 & 0.182 \\
PATN \cite{zhu2019progressive}  & 0.677 & 2.779 & 0.996 & 0.64 & 1.000 & 0.294 & 0.183 \\ \hline
Real Data                       & 1.000 & 2.431 & 0.984 & 1.00 & 1.000 & 0.000 & 0.000 \\ \hline
\end{tabular}%
}
\end{table}

\subsection{Ablation}
\begin{figure}
  \centering
  \includegraphics[width=\linewidth]{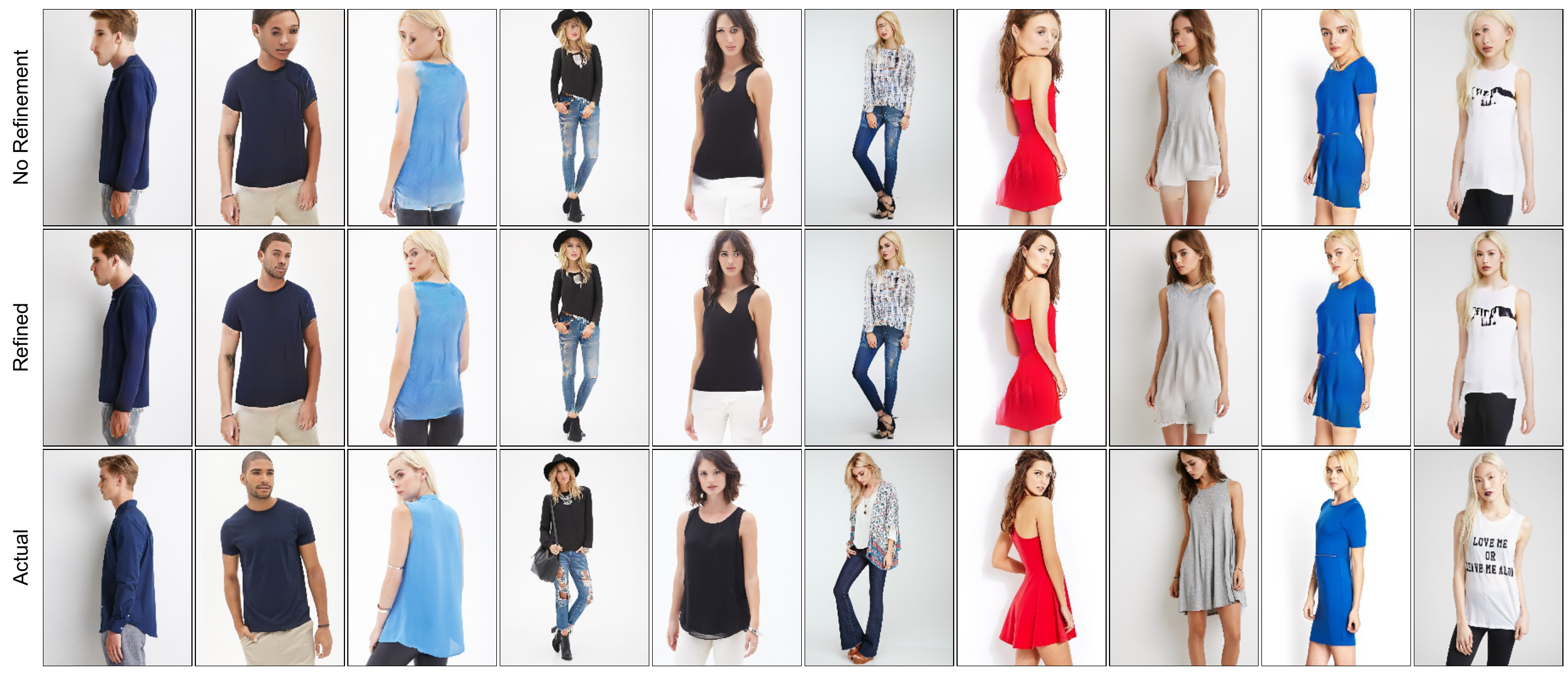}
  \caption{Qualitative results by the proposed pipeline using the partially text-guided generator with and without refinement.}
  \label{fig:result_stage2b}
\end{figure}

We perform exhaustive ablation experiments to understand the effectiveness of different architectural components of the proposed pipeline. As shown in Table \ref{tab:ablation_2}, refinement helps to improve SSIM, IS, and GCR scores in both partially and fully text-based approaches. Though the improvement in terms of metric values may look incremental, as shown in Fig. \ref{fig:result_stage2b}, the qualitative improvement due to the refinement operation is remarkable. Facial features play an essential role in the overall human appearance. Thus, the use of refinement is highly desirable in the pipeline. We report the rest of the ablation results with a partially text-based scheme while keeping the refinement operation intact in the pipeline.\\
We also explore several text embedding techniques and their effects on the generation pipeline. As shown in Table \ref{tab:ablation_3}, the encoding methods like FastText \cite{athiwaratkun2018probabilistic} and Word2Vec \cite{mikolov2013efficient} perform closely to BERT \cite{devlin2018bert}. Thus, we can conclude that our method is robust to standard text embedding algorithms.\\
We also observe the effect of multi-resolution attention used in $G_S$. In the case of single-scale attention, we only take point-wise multiplication of the channels of the final pose encoder and the final image encoder and skip all the following supervision of the pose encoders at higher resolution levels. As shown in Table \ref{tab:ablation_4}, multi-scale attention significantly improves majority of the evaluation metrics.

\begin{table}
\centering
\caption{Effects of source encoding and regressive refinement.}
\label{tab:ablation_2}
\resizebox{1.0\linewidth}{!}{%
\begin{tabular}{lc|ccccccc}
\hline
\multicolumn{1}{l|}{\textbf{Source Encoding}} &
  \textbf{Refinement} &
  \textbf{SSIM} &
  \textbf{IS} &
  \textbf{DS} &
  \textbf{PCKh} &
  \textbf{GCR} &
  \textbf{\begin{tabular}[c]{@{}c@{}}LPIPS\\ (VGG)\end{tabular}} &
  \textbf{\begin{tabular}[c]{@{}c@{}}LPIPS\\ (SqzNet)\end{tabular}} \\ \hline
\multicolumn{1}{l|}{Keypoints}      & \ding{56} & 0.545 & 3.221 & 0.952 & 0.53 & 0.960  & 0.404 & 0.290 \\
\multicolumn{1}{l|}{Keypoints}      & \ding{52} & 0.549 & 3.269 & 0.950 & 0.53 & 0.963  & 0.402 & 0.290 \\
\multicolumn{1}{l|}{Text Embedding} & \ding{56} & 0.545 & 3.261 & 0.952 & 0.53 & 0.960  & 0.404 & 0.290 \\
\multicolumn{1}{l|}{Text Embedding} & \ding{52} & 0.549 & 3.296 & 0.950 & 0.53 & 0.963  & 0.402 & 0.289 \\ \hline
Real Data                           &           & 1.000 & 3.790 & 0.948 & 1.00 & 0.995  & 0.000 & 0.000 \\ \hline
\end{tabular}%
}
\end{table}

\begin{table}
\centering
\caption{Effects of different text encoding methods.}
\label{tab:ablation_3}
\resizebox{1.0\linewidth}{!}{%
\begin{tabular}{l|ccccccc}
\hline
\textbf{Text Embedding} &
  \textbf{SSIM} &
  \textbf{IS} &
  \textbf{DS} &
  \textbf{PCKh} &
  \textbf{GCR} &
  \textbf{\begin{tabular}[c]{@{}c@{}}LPIPS\\ (VGG)\end{tabular}} &
  \textbf{\begin{tabular}[c]{@{}c@{}}LPIPS\\ (SqzNet)\end{tabular}} \\ \hline
Multi-hot                                     & 0.558 & 3.228 & 0.953 & 0.60 & 0.970 & 0.388 & 0.274 \\
BERT \cite{devlin2018bert}                    & 0.549 & 3.269 & 0.950 & 0.53 & 0.963 & 0.402 & 0.290 \\
FastText \cite{athiwaratkun2018probabilistic} & 0.548 & 3.275 & 0.949 & 0.52 & 0.968 & 0.399 & 0.285 \\
Word2Vec \cite{mikolov2013efficient}          & 0.550 & 3.251 & 0.949 & 0.52 & 0.973 & 0.401 & 0.289 \\ \hline
Real Data & 1.000 & 3.790 & 0.948 & 1.00 & 0.995 & 0.000 & 0.000 \\ \hline
\end{tabular}%
}
\end{table}

\begin{table}
\centering
\caption{Effects of multi-resolution attention.}
\label{tab:ablation_4}
\resizebox{1.0\linewidth}{!}{%
\begin{tabular}{l|ccccccc}
\hline
\textbf{Pose Transfer Method} &
  \textbf{SSIM} &
  \textbf{IS} &
  \textbf{DS} &
  \textbf{PCKh} &
  \textbf{GCR} &
  \textbf{\begin{tabular}[c]{@{}c@{}}LPIPS\\ (VGG)\end{tabular}} &
  \textbf{\begin{tabular}[c]{@{}c@{}}LPIPS\\ (SqzNet)\end{tabular}} \\ \hline
Single-scale Attention Guided & 0.540 & 3.170 & 0.921 & 0.54 & 0.954 & 0.415 & 0.298 \\
Multi-scale Attention Guided  & 0.549 & 3.269 & 0.950 & 0.53 & 0.963 & 0.402 & 0.290 \\ \hline
Real Data                     & 1.000 & 3.790 & 0.948 & 1.00 & 0.995 & 0.000 & 0.000 \\ \hline
\end{tabular}%
}
\end{table}

\section{Limitations}\label{sec:limitation}
\begin{figure}[t]
  \centering
  \includegraphics[width=\linewidth]{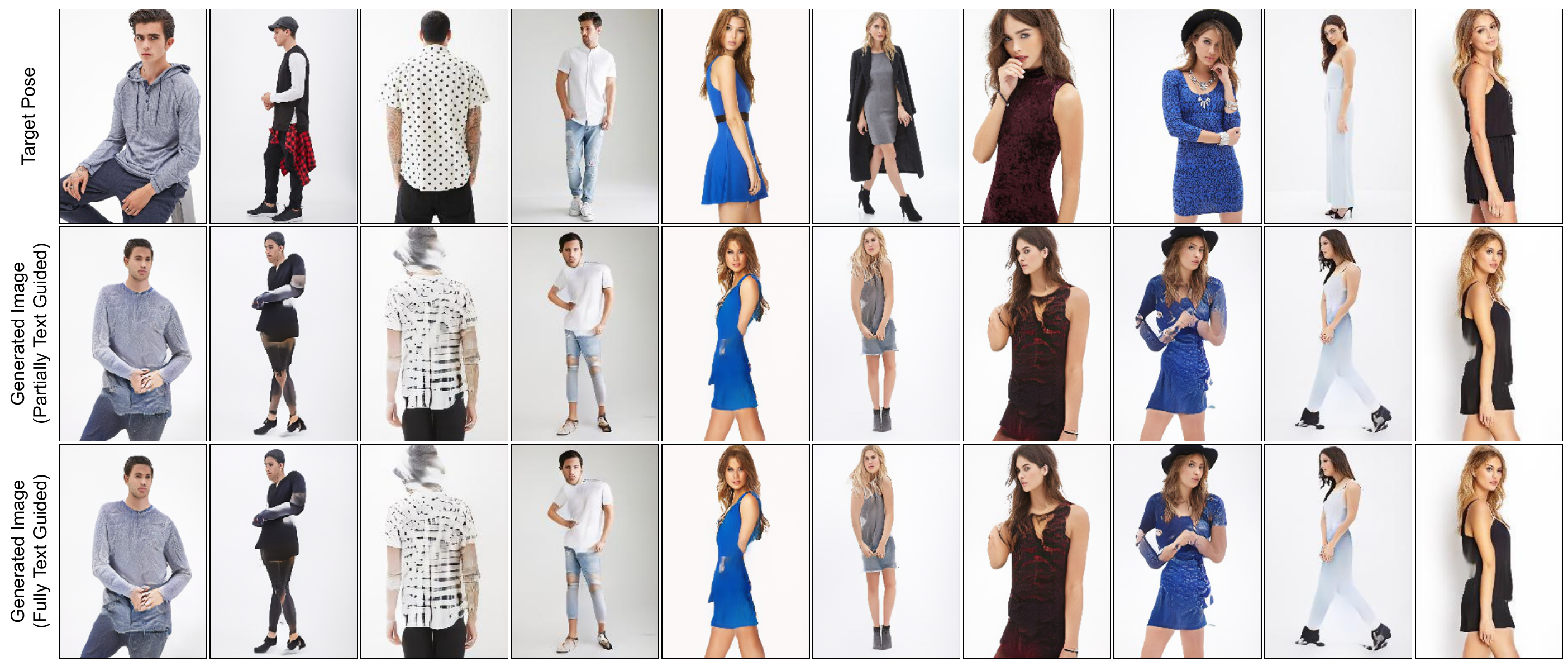}
  \caption{Failure cases of the proposed framework.}
  \label{fig:result_failure}
\end{figure}

To the best of our knowledge, this is one of the earliest attempts to transfer pose using textual supervision. As shown in Sec. \ref{sec:results}, although the results produced by the proposed approach are often at par with the existing keypoint-based baseline, it fails to perform well in some instances. When the textual description is brief and lacks a fine-grained description of the pose, the generator $G_T$ fails to interpret the pose correctly. Some of the failed cases produced by our algorithm are shown in Fig. \ref{fig:result_failure}.

\section{Conclusion}\label{sec:discussion}
In this paper, we have shown that the existing keypoint-based approaches for human pose transfer suffer from a significant flaw that occasionally prevents these techniques from being useful in real-world situations when the target pose reference is unavailable. Thus, we propose a novel text-guided pose transfer pipeline to mitigate the dependency on the target pose reference. To perform the task, first, we have designed a \emph{text to keypoints} generator for estimating the keypoints from a text description of the target pose. Next, we use a linear \emph{refinement} network to regressively obtain a refined spatial estimation of the keypoints representing the target pose. Lastly, we \emph{render} the target pose by conditioning a multi-resolution attention-based generator on the appearance of the source image. Due to the lack of similar public datasets, we have also introduced a new dataset DF-PASS, by extending the DeepFashion dataset with human annotations for poses.

\section*{Acknowledgment}\label{sec:acknowledgment}
This work was partially supported by the Technology Innovation Hub, Indian Statistical Institute Kolkata, India. The ISI-UTS Joint Research Cluster (JRC) partly funded the project.

%
%
\bibliographystyle{splncs04}
\bibliography{references}

\clearpage
\renewcommand{\thefigure}{S\arabic{figure}}
\setcounter{figure}{0}

\section*{\centering \LARGE \textbf{Supplementary Material}}

\begin{figure}
\centering
\includegraphics[width=\textwidth]{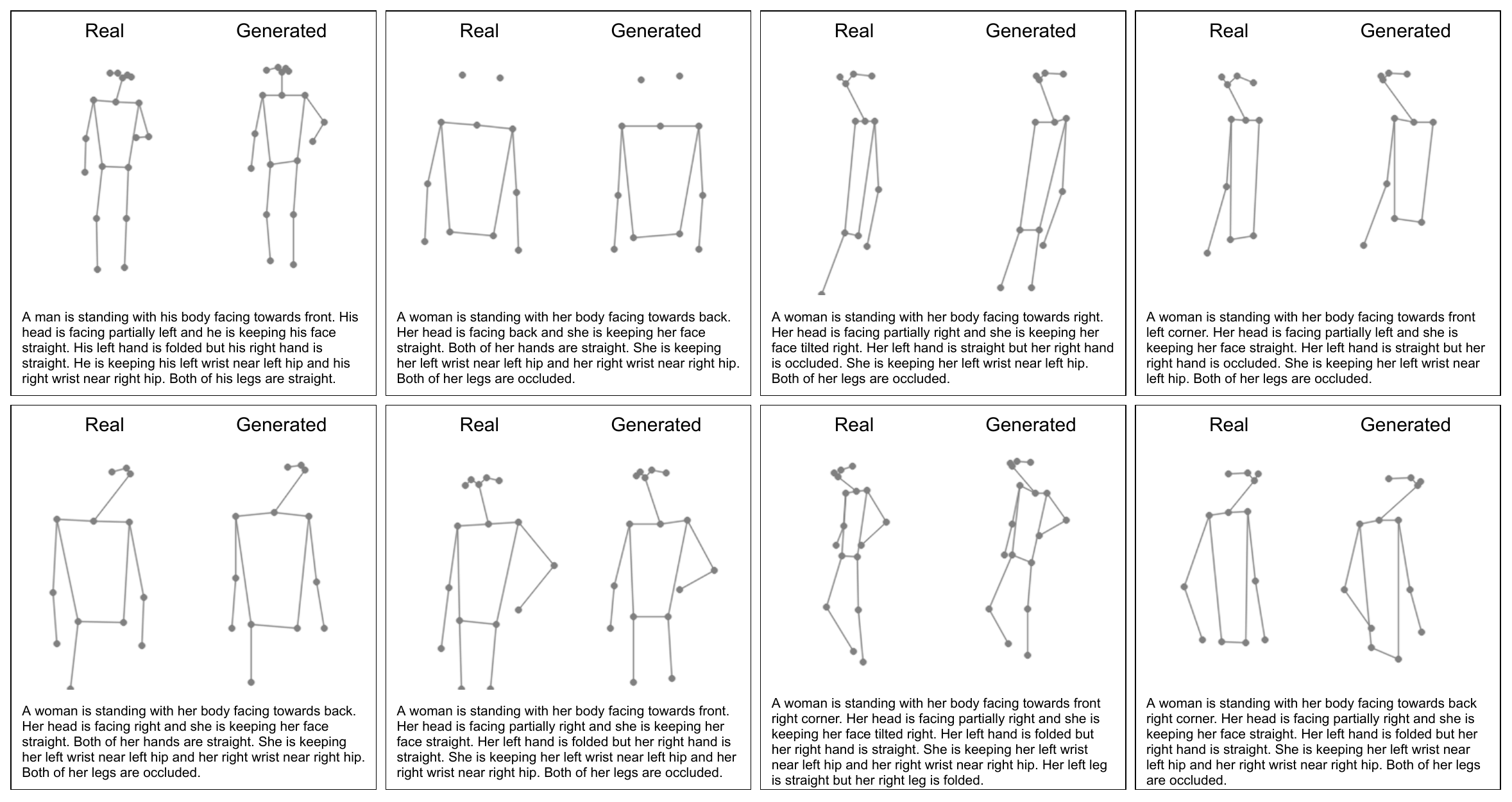}
\caption{Additional qualitative results of text to pose generation in stage 1. For each example, \textbf{Left:} Actual target pose (Ground Truth), \textbf{Right:} Generated pose conditioned purely on the textual description of the target pose, \textbf{Bottom:} Textual description of the target pose.}
\label{fig:text2pose}
\end{figure}

\begin{figure}
\centering
\includegraphics[width=\textwidth]{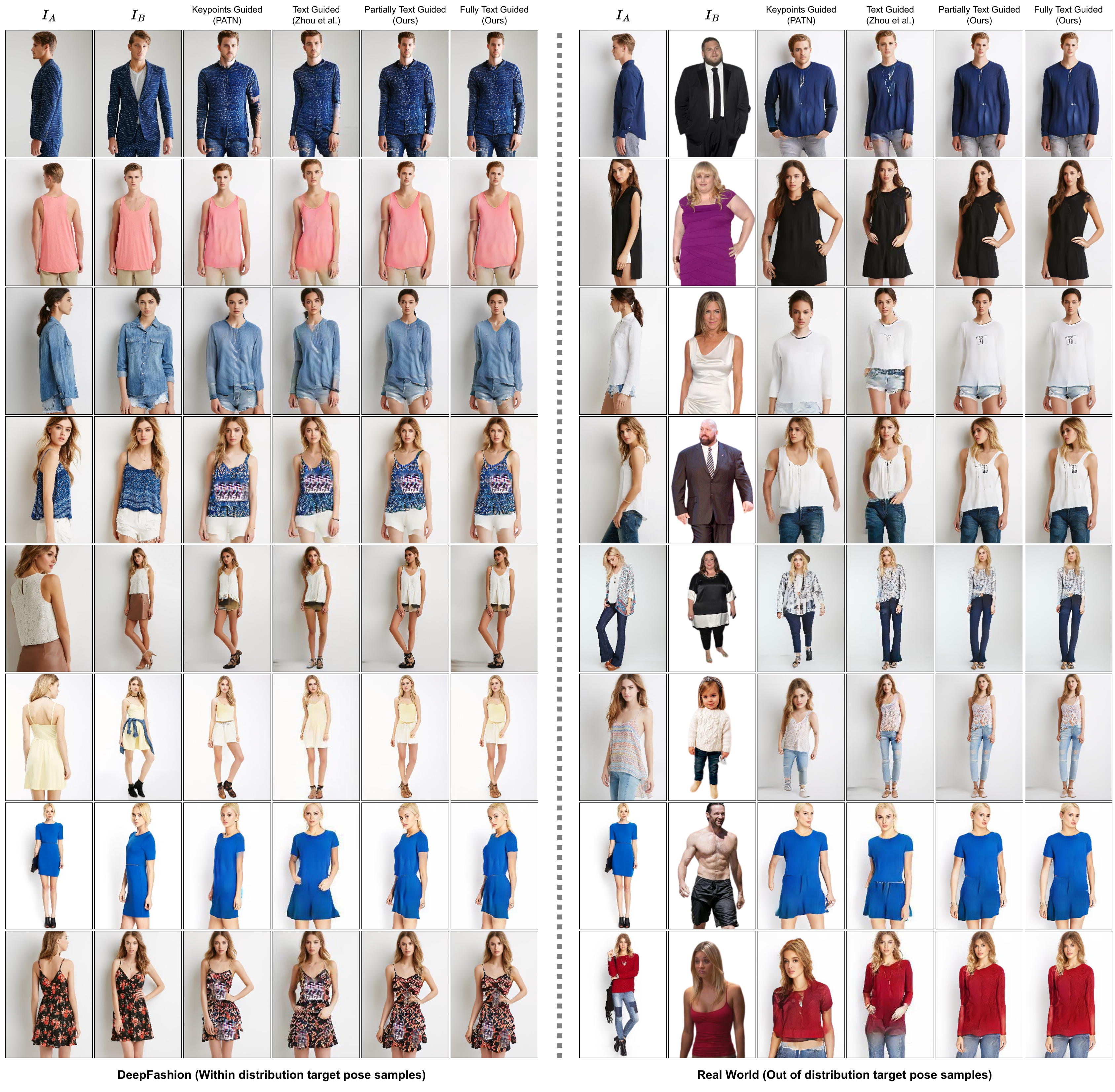}
\caption{Additional qualitative comparison among different pose transfer algorithms. Keypoint-guided methods tend to produce structurally inaccurate results when the physical appearance of the target pose reference significantly differs from the condition image. This observation is more frequent for the \emph{out of distribution} target poses. The proposed text-guided technique successfully addresses this issue while retaining the ability to generate visually decent results close to the keypoint-guided baseline.}
\label{fig:pipeline}
\end{figure}

\begin{figure}
\centering
\includegraphics[width=\textwidth]{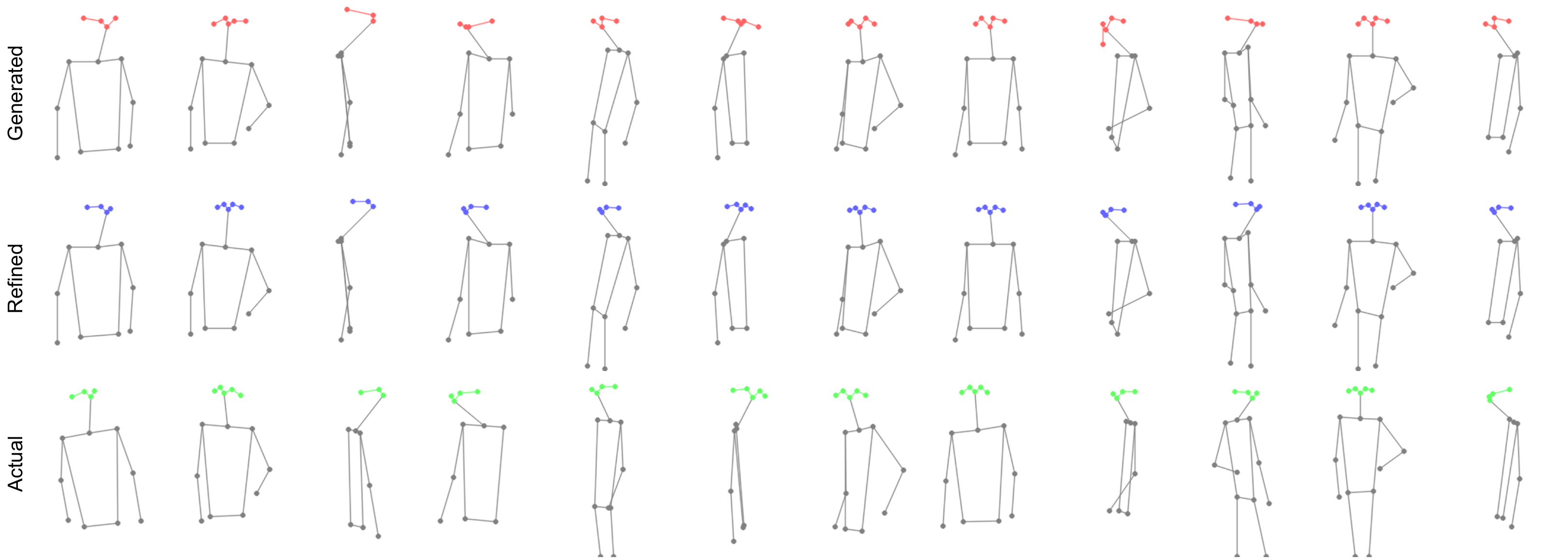}
\caption{Additional qualitative results of regressive refinement in stage 2. The refinement is performed specifically on the facial keypoints (marked with color). \textbf{Top:} Estimated keypoints from textual description in stage 1. \textbf{Middle:} Refined keypoints in stage 2. \textbf{Bottom:} Actual keypoints (Ground Truth).}
\label{fig:refinenet}
\end{figure}

\begin{figure}
\centering
\includegraphics[width=\textwidth]{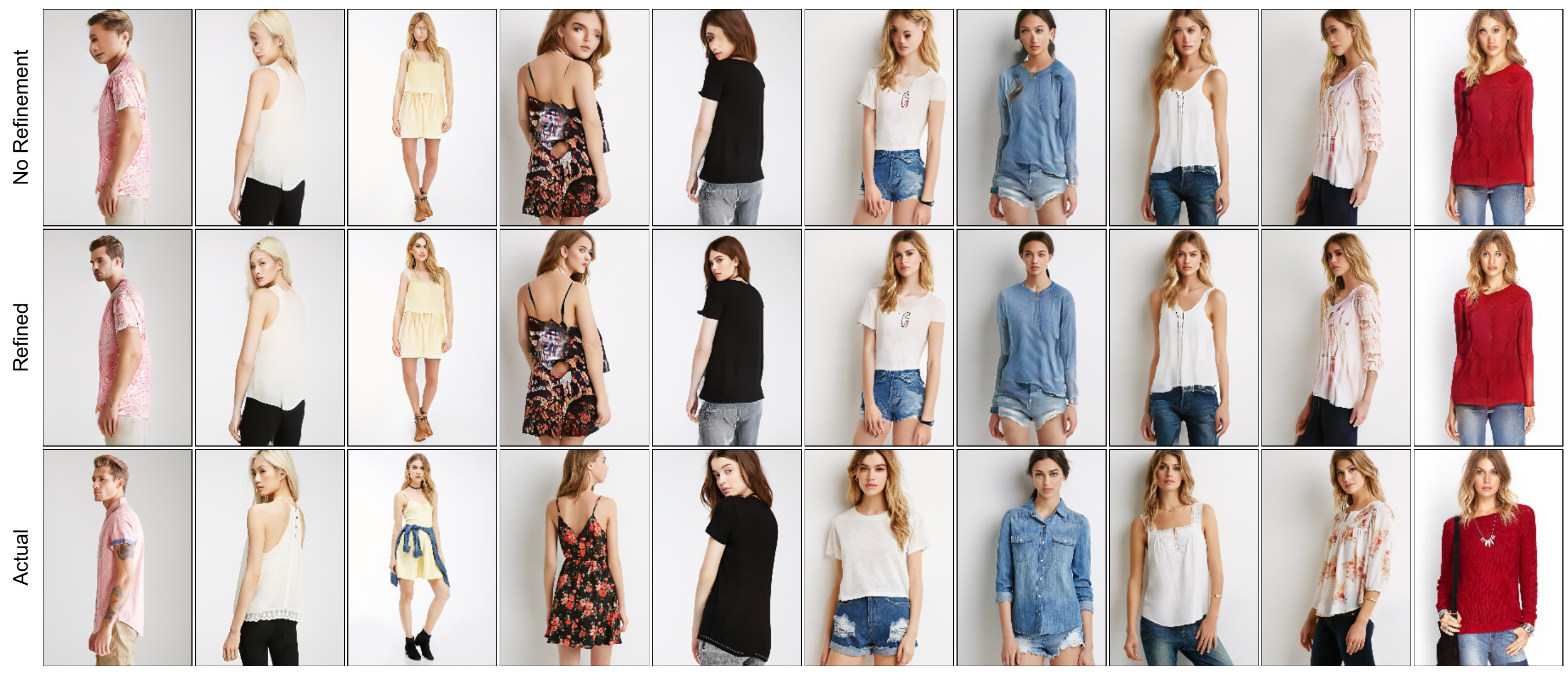}
\caption{Additional qualitative results with and without refinement. \textbf{Top:} Images generated without refinement. \textbf{Middle:} Images generated with refinement. \textbf{Bottom}: Actual target images (Ground Truth). The regressive refinement (stage 2) significantly improves the final generation quality by correcting the spatial coordinates of the keypoints estimated from textual description (stage 1).}
\label{fig:refinement}
\end{figure}

\begin{figure}
\centering
\includegraphics[width=\textwidth]{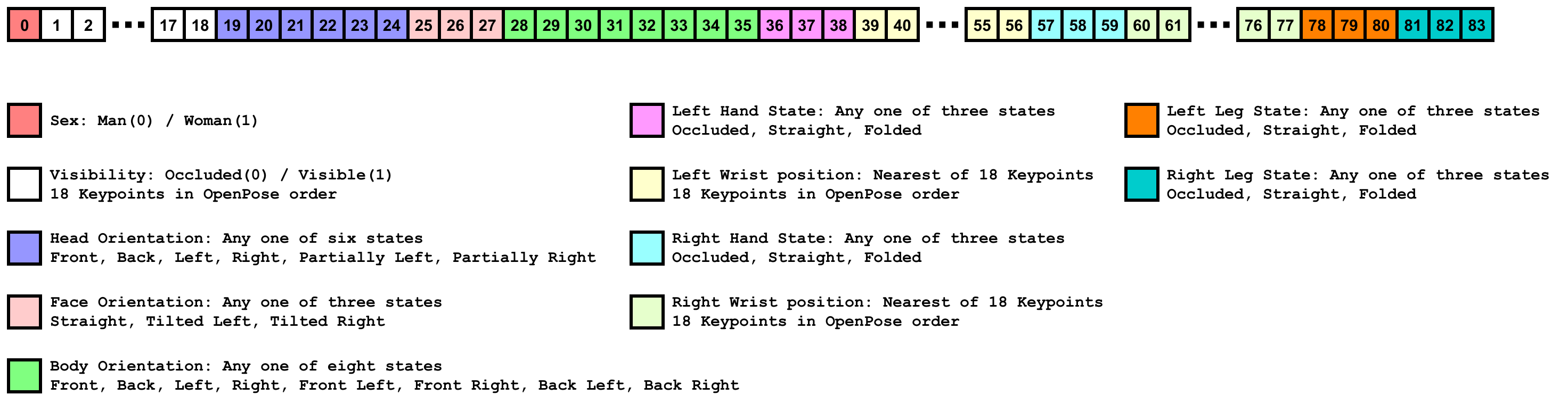}
\caption{The layout of the many-hot encoding vector in the proposed DF-PASS dataset.}
\label{fig:encoding}
\end{figure}

\begin{figure}
\centering
\includegraphics[width=\textwidth]{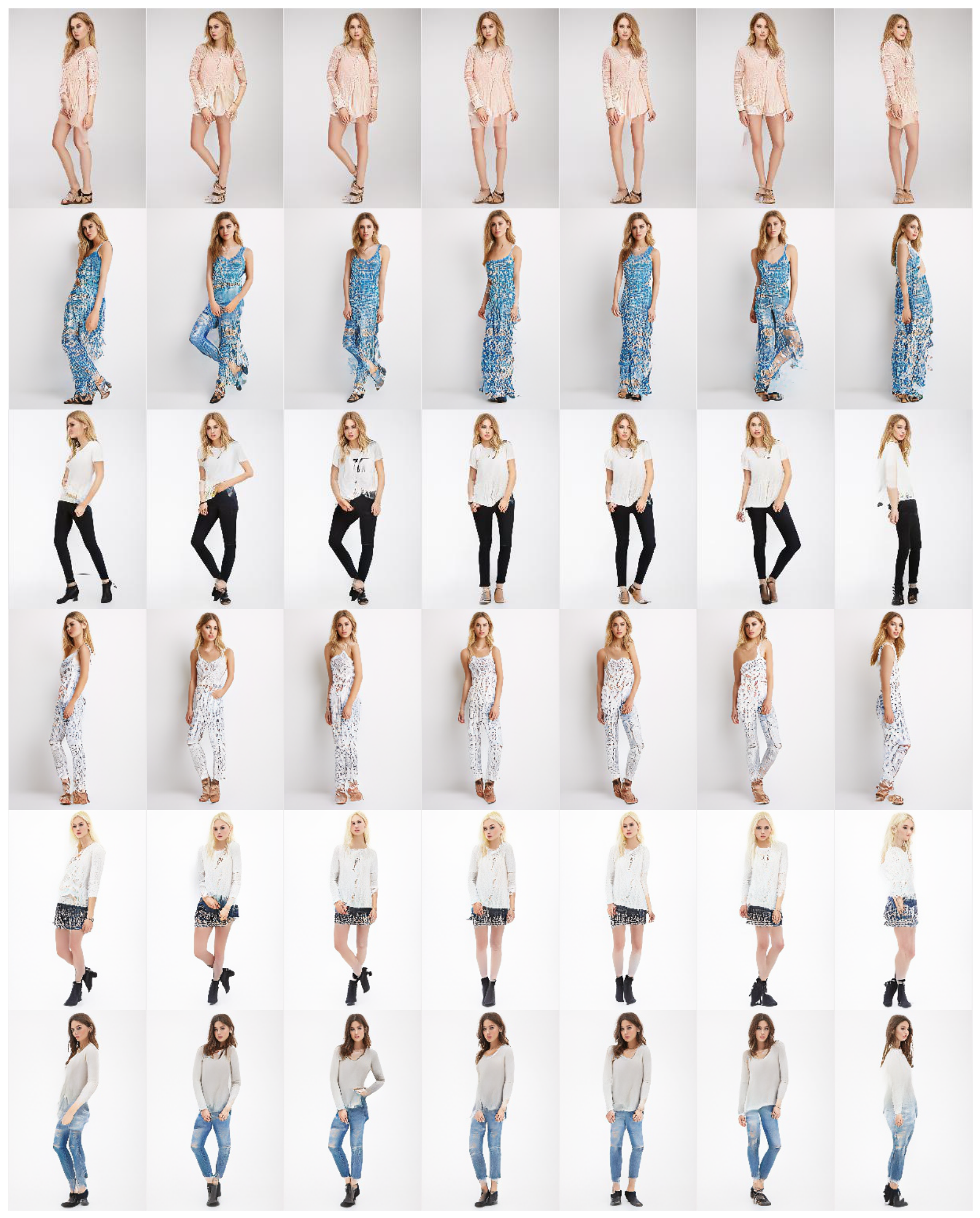}
\caption{Text-assisted $180^\circ$ interpolation in standing pose using the proposed method.}
\label{fig:interpolation}
\end{figure}

\end{document}